\documentclass{bmvc2k}

\usepackage{algorithm}
\usepackage{authblk}
\usepackage[noend]{algpseudocode}

\title{Hide and Seek tracker: Real-time recovery from target loss}

\addauthor{Alessandro Bay$^*$}{alessandro.bay@cortexica.com}{1}
\addauthor{Panagiotis Sidiropoulos$^{1,}$}{p.sidiropoulos@ucl.ac.uk}{2}
\addauthor{Eduard Vazquez}{eduard.vazquez@cortexica.com}{1}
\addauthor{Michele Sasdelli$^*$}{michele.sasdelli@gmail.com}{1}

\addinstitution{
 Cortexica Vision Systems\\
 30, Stamford Street\\
 London, UK 
}
\addinstitution{
 Mullard Space Science Laboratory\\
 University College London\\
 Holmbury St.\ Mary Dorking, UK 
 \\ \vspace{0.6cm} \hspace{-2mm}$^*$\scriptsize{The two authors contributed equally}
}

\runninghead{A. Bay, P. Sidiropoulos, E. Vazquez, M. Sasdelli }{Hide and Seek tracker}

\begin{document}

\maketitle

\begin{abstract}

In this paper, we examine the real-time recovery of a video tracker from a target loss, using information that is already available from the original tracker and without a significant computational overhead. More specifically, before using the tracker output to update the target position we estimate the detection confidence. In the case of a low confidence, the position update is rejected and the tracker passes to a single-frame failure mode, during which the patch low-level visual content is used to swiftly update the object position, before recovering from the target loss in the next frame. Orthogonally to this improvement, we further enhance the running average method used for creating the query model in tracking-through-similarity. The experimental evidence provided by evaluation on standard tracking datasets (OTB-50, OTB-100 and OTB-2013) validate that target recovery can be successfully achieved without compromising the real-time update of the target position.

\end{abstract}

\section{Introduction}\label{sec:intro}

Recent advances in deep learning architectures significantly improved the state of the art in a number of computer vision applications. In general, it is rather common in computer vision applications to rely on a convolutional neural network (CNN) pre-trained on large datasets such as ImageNet \cite{krizhevsky2012imagenet}, which is subsequently fine-tuned to be transferred to a different task. Examples include the use of 
CNNs for object detection \cite{girshick2014rich}, semantic segmentation \cite{long2015fully}, caption generation \cite{xu2015show}, action recognition \cite{simonyan2014two}, etc.

On the other hand, the task of visual tracking consists of the ability of tracking arbitrary objects in videos, given the starting position in the initial frame \cite{AYilmaz06}. The inherent ad-hoc nature of this setup has caused that deep-learning based architectures struggled to show disruptive changes in tracking for some years \cite{HNam16}.
However, recently, major advances in deep-learning architectures applicable to tracking has significantly improved the state-of-the-art, both for real-time \cite{bertinetto2016fully,valmadre2017end} and non-real-time tracking \cite{SYun17,MDanelljan16}. 

A significant adjustment towards this path was to gradually replace the transfer learning from large classification datasets with techniques that either perform on-line training (e.g. ECO \cite{MDanelljan17}) or are trained off-line in a setup that is only loosely connected to semantic classification (e.g. CFNet \cite{valmadre2017end}).
The on-line methods currently achieve state-of-the-art performance but not real-time tracking, while the off-line methods the opposite.
While both on-line and off-line techniques progressively converge to the optimisation of the tracker accuracy and its computational complexity, several significant challenges remain unresolved, including the sensitivity to distractors \cite{LWang15}, object disappearance and re-appearance events \cite{JValmadre18} and the recovery from target loss \cite{RTao17}.

This work deals with the latter, i.e. the real-time recovery of a tracker from a temporary failure. The difficulty of tracker recovery originates from its main design principle, i.e. its ability to accumulate correct object positions for a substantial amount of time. As a matter of fact, tracking could be viewed as an accumulative retrieval problem, in which the performance is evaluated based on the algorithm potential to retrieve the correct object position in each and every frame that constitute a video signal. This potential is reversed to a significant flaw in the cases that the tracker would temporarily lose the position of the object due to an abrupt camera movement, an unexpected and abrupt object movement, a technical problem, etc.
Once the sampling drift \cite{RTao17} causes the tracker bounding box to not intersect with the object, the tracking algorithm potential ``secures'' that it will remain in the background with little possibility of recovery.

A possible method to tackle this issue is to exploit the tracking-through-similarity para-digm that is adopted by CFNet \cite{valmadre2017end}. The assumption that a similarity search among candidates is characterised as ambiguous/unambiguous depending on the global maximum magnitude in relation to the local maxima is rather common in computer vision (e.g. in image matching \cite{DLowe04}). Following a similar rationale, it is suggested to declare ambiguous a tracking-through-similarity tracker output, in the cases that the similarity peak is not prominent. Subsequently, the tracker could pass on a ``failure mode'', during which the target would recover from the possible target loss before returning to ``normal mode''.
In this work we introduce the ``Hide and Seek'' (HnS) architecture and we experimentally tested it on state-of-the-art benchmarks.

Additionally, the original CFNet algorithm is using a sub-optimal approach to estimate the running average of the feature map used for query. More specifically, the CFNet running average is updated very slowly, thus causing a dependency on the object visibility on the first video frame. This flaw is not striking in the OPE evaluation because for most of the benchmark videos the object to be tracked is clearly visible in the first frame. However, this introduces an unnecessary sensitivity to the initialisation frame (apparent in the TRE evaluation), which could be avoided if the running average estimation becomes more smooth. 

In summary, apart from the original architecture, the main contributions of the paper are the following:
\begin{enumerate}
\item The ambiguity measure that estimates the confidence on the tracker output.
\item The enlargement of the search area when the tracker is on ``failure mode''.
\item The use of a simple low-level representation as a short-term, backup tracker.
\item An improved running average method for the query model of the tracker.
\end{enumerate}

The paper is structured as follows.
The relevant literature is reviewed in Section \ref{sec:relworks}, before passing to the presentation of the HnS tracker in Section \ref{sec:HnS}.
The experimental evaluation is conducted in Section \ref{sec:res}, while Section \ref{sec:concl} concludes this work.

\section{Related works}\label{sec:relworks}
\subsection{Real-time deep learning trackers}

Due to its major significance, the tracking of moving objects in videos has a long history of research and development. A qualitative comparison of a review on early-days trackers \cite{AYilmaz06} and a corresponding publication almost a decade later \cite{AWMSmeulders14} (the later just before deep-learning trackers) reveals the substantial progress achieved in the meantime, which has led to algorithms that could perform tracking with high precision \cite{SHare11}, in real-time \cite{DSBolme10} and with the additional capability to recover from severe occlusions or the loss of target \cite{ZKalal10}.

The maturity of the domain was one of the reasons that deep-learning trackers were initially struggling to outperform ``classical'' trackers \cite{HNam16}. But perhaps a more important reason was the inherent characteristics of the tracking setup that undermined a straightforward transfer of deep-learning techniques to this task \cite{DZhang17}. More specifically, (1) maximising heatmaps corresponding to semantic classes is not necessarily the optimal strategy to locate a specific object \cite{CMa15} (especially when distractors are present), (2) off-line training is hampered by the lack of large datasets for this task, and (3) on-line training is prohibitively slow for applications requiring real-time tracking, especially if the model update is conducted in each and every frame \cite{MDanelljan17}, \cite{RTao17}.

On the other hand, the visual similarity of the tracked object between two consecutive frames is implied by the small temporal window between them ($0.04$ secs for a $25$fps video). Based on this rationale, Bertinetto et al. introduced a tracking-by-similarity tracker \cite{bertinetto2016fully} that despite its simple architecture achieved state-of-the-art real-time performance. In this setup, the similarity is learned through a Siamese deep network that is trained offline, while the localisation is conducted through a correlation filter \cite{valmadre2017end}. 

One of the main issue with such a tracker is the sampling drift \cite{RTao17} (from which the tracker often fails to recover) that occurs in the case of abrupt object or/and camera motion.
Two main solutions have been recently proposed: (1) in \cite{JValmadre18} the authors assume that the heatmap that is generated in the final stage of the algorithm incorporates the localisation ambiguity by not exhibiting a clear peak and it temporarily discards peaks below a threshold, and (2) in \cite{RTao17} it is suggested to interleave global similarity search (in the whole frame) every $N$ frames ($N$ being a hard-coded constant parameter) to avoid the model drift.

In this work, we carefully amalgamate the above solutions in order to optimise both the accuracy and the computational time.
Firstly, due to the parameter sensitivity of a comparison to a hard-coded value, the  localisation ambiguity is detected using the Nearest Neighbour Distance Ratio (NNDR).
If a localisation is declared ambiguous the tracker passes to ``failure mode'', during which the window is expanded (but without covering the whole frame, as in \cite{RTao17} so as to reduce the computational cost). Finally, while the tracker is in failure mode, tracking is conducted using correlation of low-level patch representations.
The latter builds upon the recent improvements on tracker performance that was achieved by including the first network layers, which model the low-level image content \cite{MDanelljan16}, \cite{MDanelljan17}.
In the current work, the low-level image content is modelled though Census transform \cite{HHirschmuller09}, in order to not compromise the computational cost, and because Census transform has exhibited exceptional robustness in low-level matching \cite{HHirschmuller09}. 


\subsection{Nearest Neighbour Distance Ratio}

The use of Nearest Neighbour Distance Ratio (NNDR), i.e. the ratio of the distance to the nearest neighbour over the distance to the second nearest neighbour, to validate candidate matches is an idea originating from computer vision \cite{DLowe04} and has been extensively used in applications such as image matching (e.g.\cite{SPaul16}, \cite{ASedaghat15}, \cite{PSidiropoulos15}) to discard erroneous results. 

The basic assumption of NNDR is that the value corresponding to the correct estimation and the values corresponding to all false estimations derive from two distinct and non-overlapping distributions.
Following this assumption, the two nearest neighbour values should be samples from separate distributions, thus generating a high NNDR value.
Conversely, a low NNDR value implies that the two nearest neighbours are sampled from the same distribution, therefore, the detection result may be declared ambiguous and discarded.

The most important property of NNDR, which is in large the root cause of its extensive use in the literature, is its performance robustness.
NNDR is expected to exhibit almost identical performance for a large range of threshold values \cite{PSidiropoulos15}, thus reducing the parameter sensitivity of the algorithm employed.
For a similar reasons, thresholding the nearest neighbour value is a sub-optimal approach in applications in which the nearest neighbour values exhibit significant variance and their magnitude is difficult to be systematically predicted. 

Due to the NNDR robustness and high performance, in this work we propose using it to discriminate between ambiguous and unambiguous peaks in the tracker heatmap. While the application setup is different, from a data-science point of view the similarity is apparent.
The tracker heatmap is a matrix of distances, one of which corresponds to the actual object position while the rest corresponds to erroneous entries that can be in general assumed to be selected from a distinct distribution.
The use of NNDR could allow the identification of ambiguous peaks, thus allowing the pipeline to pass to ``failure mode'', before recovering the object position in subsequent frames and continuing in its ``normal mode''.

\section{Hide and seek (HnS) tracker}\label{sec:HnS}
\begin{figure}[h]
\centering{
\includegraphics [scale=0.55] {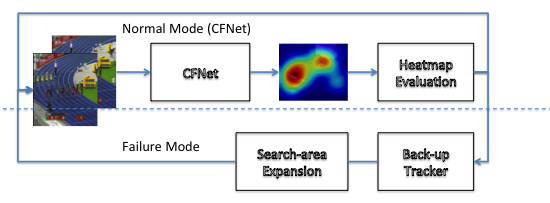}
\caption{The architecture of the HnS tracker}
\label{fig:architecure}}
\end{figure}

In this Section we describe Hide and Seek (HnS, Figure \ref{fig:architecure}), our novel tracker that builds upon CFNet in order to achieve real-time target recovery.

\subsection{Heatmap confidence evaluation}\label{subsec:heatmap_eval}
The main concept of our architecture lies in the assumption that the CFNet heatmap output can be used to evaluate the confidence on the bounding box update.
The confidence is modelled through the ratio of the two most dominant peaks, which are estimated as follows: first the correlation filter map is projected onto $yz$ and $xz$ planes, before being differentiated twice in order to identify the local maxima, which determine the two most dominant peaks.
Since CFNet is estimating patch similarity, these peaks correspond to the nearest neighbour and the second nearest neighbour of the current bounding box. Their ratio (i.e. NNDR) is used to evaluate the confidence on the tracker output.

More specifically, if the ratio is above the confidence threshold the tracker output is considered safe, therefore, the top peak is followed to update the object position and the tracking continues following the CFNet algorithm in the next frame. On the other hand, if the ratio is below the confidence threshold the tracker output is considered ambiguous and the tracker passes on failure mode.

The following measures are taken during the time that the tracker is on failure mode: (1) CFNet output is not used to update the object position (the top peak position is ignored), (2) the object position is updated following the estimation of the backup tracker, and (3) in the following frame 
the object is searched in an area wider that the original one.

The backup tracker is based on correlating Census-transformed \cite{RZabih94} image patches.
Census transform is a simple and powerful low-level representation of the image content that presents a set of positive characteristics: (1) it has linear computational complexity, (2) it preserves the object edges, (3) it is robust to radiometric differences (which may occur during tracking due to abrupt camera motion) \cite{HHirschmuller09}, and (4) it generates robust optic flow estimations \cite{DHafner13}.
The 8-bit binary strings that Census transform generates for each pixel are converted into $4$ decimal numbers by iteratively applying a circular shift of $2$ positions before conversion.
The result is correlated and the position of the maximum value is followed to update the object bounding box.
It should be noted that from the deep-learning point of view the Census transform could be considered as a hand-crafted filter of a single layer of a neural network.

\subsection{Smooth running average}\label{subsec:smooth_average}
The strategy used in CFNet \cite{bertinetto2016fully} to create the query model from the previously seen feature map is a simple running average:
\begin{equation} \label{Eq:1}
Q_1=F_1; ~~\ \ Q_n= Q_{n-1}(1-\alpha) + \alpha F_n
\end{equation}
where $Q_n$ is the $n$-th query model and $F_n$ is the $n$-th feature map and the update factor $\alpha$ was empirically set to $\alpha=0.005$.
As a result, during the first video frames (in which case $n \not\gg 1/\alpha$), $Q_n$ is dominated by the first feature map $F_1$. 

This approach is a reasonable measure against sampling drift in OPE benchmarking, since the first frame of the video usually captures the object from an angle that allows a clear identification (e.g. a person would be usually captured from an angle that makes the face fully visible and not from the back).
However, such a strong dependence from $F_1$ is expected to be suboptimal in the general case, especially in practical applications that the tracked object would be initialised with an arbitrary view and with sub-optimal quality.

The dependence from the first frame is reduced by updating Eq. \ref{Eq:1} as follows:
\begin{equation} \label{Eq:2}
Q_1 = F_1 ; \ \  Q_n = Q_{n-1}(1 - 0.5/n - \alpha) + (0.5/n + \alpha)F_n
\end{equation}
Eqs. \ref{Eq:1} and \ref{Eq:2} converge asymptotically. Their main difference is that Eq. \ref{Eq:2} generates a ``smooth average'' over the first frames, by creating a ``bootstrap'' model which uses a significant number of initial frames ($\sim \frac{1}{5\alpha}$), instead of a single frame.
As it will be demonstrated in the next Section, this improves significantly the performance in the more challenging TRE evaluation, a benchmark that measures the robustness of the tracker on the initial object view. 

The introduced algorithm is analytically presented in Algorithm \ref{alg:hideandseek}, where we highlight in bold the differences from the classic CFNet algorithm. The parameter values are the ones used in our implementation.

\begin{algorithm} \footnotesize
\caption{Hide and seek algorithm (HnS)}\label{alg:hideandseek}
\begin{algorithmic}[1]
\Procedure{HnS tracker}{$video,b_0$} \Comment{It returns the track of $video$ starting from box $b_0$}

\State $\bf{ LARGE\_SEARCH  \gets false }$ \

\State $\bf{ FAILURE\_MODE  \gets false }$ \

\State $crop  \gets crop\_image( video[0], b_0) $ \

\State $crop  \gets resize\_for\_nn( crop, LARGE\_SEARCH ) $ \

\State $NN  \gets create\_NN( instance\_size ) $

\State $running\_average  \gets NN.forward\_pass( crop ) $

\State $ // \ initialize \ the \ box $

\State $b  \gets b_0 $

\For{$frame$ in $video$}

  \State $crop  \gets crop\_image( frame, b )$
  
  \State $crop  \gets resize\_for\_nn( crop, LARGE\_SEARCH )$
  
  \State $feature\_map  \gets NN.forward\_pass( crop )$
  
  \State $correlation\_map  \gets convolution( feature\_map, running\_average )$
  
  \State $weight\_window  \gets create\_weight\_window( hann, size(correlation\_map) ) $
  
  \State $correlation\_map  \gets correlation\_map * weight\_window $
  
  \If{ $ \bf{ height\_second\_peak( correlation\_map ) > 0.9 }$ }
     \State $\bf{b  \gets BACKUP\_TRACKER}$ 
     
     \State $ \bf{ LARGE\_SEARCH  \gets true }$
     
     \If{ $ \bf{ not \ FAILURE\_MODE } $ }
     
       \State $ \bf{instance\_size  \gets 2 * instance\_size } $
       
       \State $ \bf{ NN \gets create\_NN( instance\_size ) }$
     
     \EndIf
     
     \State $ \bf{ FAILURE\_MODE \gets true }$
     
  \Else
     \If{ $ \bf{ FAILURE\_MODE } $}
     
       \State $ \bf{ instance\_size \gets instance\_size/2 }$
       
       \State $ \bf{ NN  \gets create\_NN\_with\_new\_shape( instance\_size ) } $ 
     
     \EndIf

     \State $ // \ update \ the \ position$
     \State $b  \gets find\_peak( correlation )$ 
     
     \State $crop  \gets crop\_image( frame, b_0) $
     
     \State $crop  \gets resize\_for\_nn( crop, LARGE\_SEARCH ) $

     \State $feature\_map  \gets NN.forward\_pass( crop )$
     
     \State $ // \ update \ the \ running \ average \ of \ the \ model$
  \State $ \bf{running\_average \gets (0.995 - 0.5/n) * running\_average  + (0.5/n) * feature\_map}$
  
  \EndIf

\EndFor\label{euclidendwhile}
\EndProcedure
\end{algorithmic}

\end{algorithm}

\section{Experimental Results}\label{sec:res}

\subsection{Implementation Details}\label{subsec:implement}

The CFNet implementation that we used is the code that was provided by the authors \cite{valmadre2017end}. In order to manifest the validity of each of the measures taken when the tracker passes to failure mode,  variations of the HnS tracker has been evaluated:
\begin{itemize}
\item HnS0, in which the position of the bounding box is not updated until the NNDR is above the confidence threshold, but the search area for the subsequent frames is double than the original.
\item HnS1, in which the new position is extrapolated by the position of the object in the past two frames using bilinear interpolation.
\item HnS, in which the algorithm presented in Section \ref{subsec:heatmap_eval} is followed (but not smooth average)
\item HnSSA, where the HnS algorithm is combined with the smooth average approach.
\end{itemize}
All of the calculations were performed with MATLAB R2017a, MatConvNet 1.0-beta25, gcc/g++-4.9, Cuda-8.0, Cudnn-5.1, on a i7-6800K CPU @ 3.40GHz $\times$ 12 workstation with 32 GB RAM and a single nVidia GeForce GTX 1080Ti graphics card.

\subsection{Benchmark}\label{subsec:benchmark}

For evaluating our algorithm, we use the object tracking benchmark (OTB, \cite{WuLimYang13}). As in \cite{valmadre2017end}, we considered three benchmarks: the entire dataset with 100 videos (OTB-100), its subset comprising 50 videos (OTB-50), and the official OTB-2013 challenge. For all of them, we compare One-Pass Evaluation (OPE) and Temporal Robustness Evaluation (TRE). The TRE metric consists in choosing 20 equispaced points per video sequence and running the tracker from each of them until the end, while OPE metric processes the video from the beginning until the end, which is equivalent to the first trial of TRE. Both metrics are evaluated in terms of overlap and precision. In particular, the precision is computed as the percentage of frames whose estimated location is within a given threshold distance of the ground truth. As a representative precision score, we use the score for the threshold of 20 pixels \cite{babenko2011robust}. On the other hand, overlap is computed as the success rate of frames whose intersection over union (IoU) of the predicted bounding box and the ground truth is larger than a given threshold. In this case too, we followed the standard literature approach (e.g. \cite{LWang15}) to report the area under curve (AUC) of each success plot to rank the tracking algorithms, instead of the success rate at a specific threshold. Finally, the tracker speed (in fps) is also reported for each method.

\subsection{Results}\label{subsec:results}
The performance achieved by the $4$ HnS variations described in Section \ref{subsec:implement} are presented in Table \ref{tab:res} and compared to the CFNet2 baseline \cite{valmadre2017end}. It should be noted that due to compatibility issues between MATLAB R2017a (used in this work) and R2015a (used in \cite{valmadre2017end}), the reported results slightly diverge from the ones reported in \cite{valmadre2017end}.

\begin{table}[h]
\scriptsize{
\centering{
\begin{tabular}{l | c | c c c c | c c c c | c c c c}
& & \multicolumn{4}{c}{OTB-2013} & \multicolumn{4}{|c|}{OTB-50} & \multicolumn{4}{c}{OTB-100} \\
& & \multicolumn{2}{c}{OPE} & \multicolumn{2}{c}{TRE} & \multicolumn{2}{|c}{OPE} & \multicolumn{2}{c|}{TRE} & \multicolumn{2}{c}{OPE} & \multicolumn{2}{c}{TRE} \\
Method & fps & IoU & prec & IoU & prec & IoU & prec & IoU & prec & IoU & prec & IoU & prec \\
\hline
CFNet2 \cite{valmadre2017end} & 71.1 & 57.0 & 74.1 & 59.9 & 76.1 & 49.4 & 64.3 & 52.9 & 69.0 & 55.7 & 71.6 & 58.2 & 73.7 \\
CFNet2+HnS0 & 64.2 & 57.4 & 74.4 & 60.2 & 76.5 & 49.1 & 64.1 & 53.0 & 69.2 & 54.6 & 70.3 & 58.0 & 73.4 \\
CFNet2+HnS1 & 64.0 & 57.9 & 75.2 & 60.2 & 76.5 & 48.9 & 63.8 & 53.0 & 69.0 & 54.8 & 69.9 & 58.1 & 73.6 \\
CFNet2+HnS  & 61.6 & \bf{60.0} & {\bf78.2} & 60.6 & 76.9 & 51.3 & 68.1 & 53.9 & 70.7 & {\bf57.3} & {\bf74.1} & 58.7 & 74.4 \\
CFNet2+HnSSA & 62.3 & 58.7 & 76.8 & {\bf61.9} & \bf{79.8} & \bf{51.5} & {\bf68.5} & \bf{55.0} & {\bf73.2} & 57.1 & 73.9 & {\bf59.9} & {\bf77.0} \\
\end{tabular}
\caption{The results of our baseline (CFNet2) along with the $4$ tested variations of HnS: HnS0, HnS1, HnS and HnS+smooth average (HnSSA). The best performance is highlighted in bold.}
\label{tab:res}}}
\end{table}

A first comment is that the relative improvement achieved by HnS is $3.5\%-6.5\%$ in precision and $2.9\%-5.2\%$ in accuracy, without significant computational overhead. The fact that both HnS and HnSSA outperformed CFNet2 in all evaluation scenarios manifests the potential of the introduced method to improve tracking through the real-time recovering from a target loss. In comparison, the ambiguous results achieved by HnS0 and HnS1 imply that approaches which instead of a ``failure mode'' (backup) tracker rely on simple bounding box updates would generally fail to exploit the identification of low-confidence heatmaps. 

The comparison between HnS and HnSSA validates the analysis conducted in Section \ref{subsec:smooth_average}. HnSSA clearly outperforms HnS in all TRE evaluations, while achieves similar performance in OPE in two out of three evaluations (OTB-2013 and OTB-100) and worse in OTB-50. This is aligned with the dependence of the TRE from the object viewing angle (in the original frame), thus signifying that in most practical applications (in which the object viewing angle in the first frame is not generally known) HnSSA should be preferred.

\begin{figure}[h]
\centering{
\includegraphics[scale=0.35]{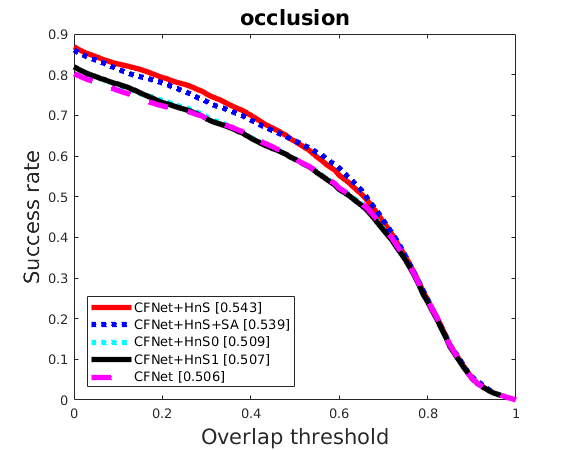}
\includegraphics[scale=0.35]{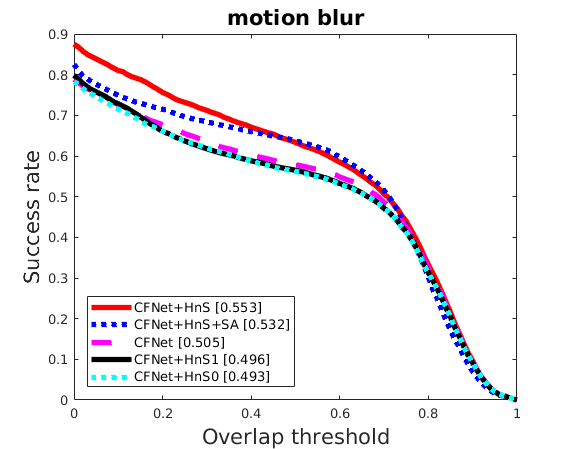}
\\
\includegraphics[scale=0.35]{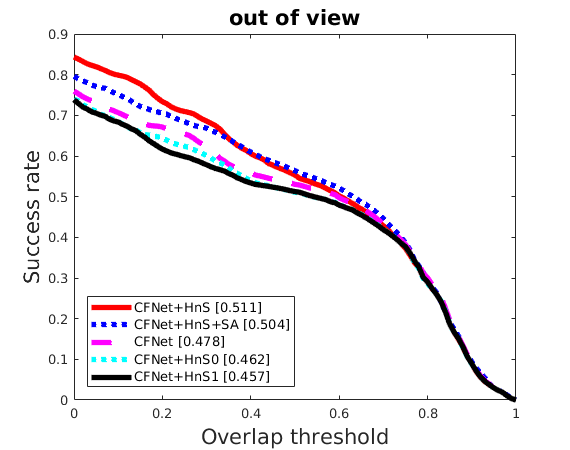}
\includegraphics[scale=0.35]{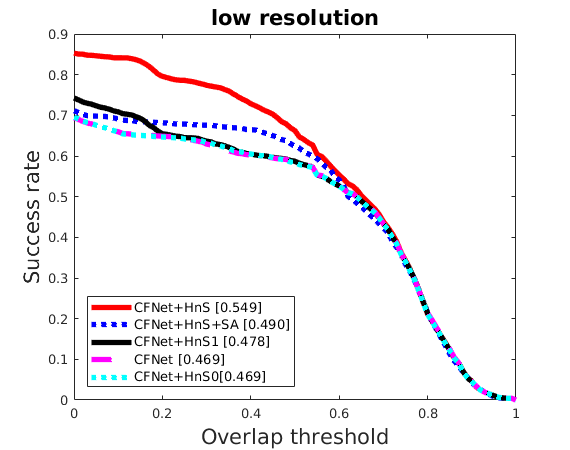}
\caption{OPE of IoU scores for relevant attributes (occlusion, motion blur, out of view, and low resolution, respectively) on the OTB-100 dataset.}
\label{fig:attr}}
\end{figure}

Analysing various attributes describing the videos, we find out that our HnS approach significantly increases CFNet performance in those videos characterised by occlusion, motion blur, out of view, and low resolution, as shown in Figure \ref{fig:attr} for OTB-100 dataset, IoU curve scores and OPE metric. Note that in the low resolution and motion blur examples, i.e. in videos of low quality, HnS clearly outperforms HnSSA.
This implies that HnS is to be preferred in applications that the video quality is expected to be low and the initial frame is of good quality.

\begin{figure}[h!]
\centering{
\includegraphics[scale=0.25]{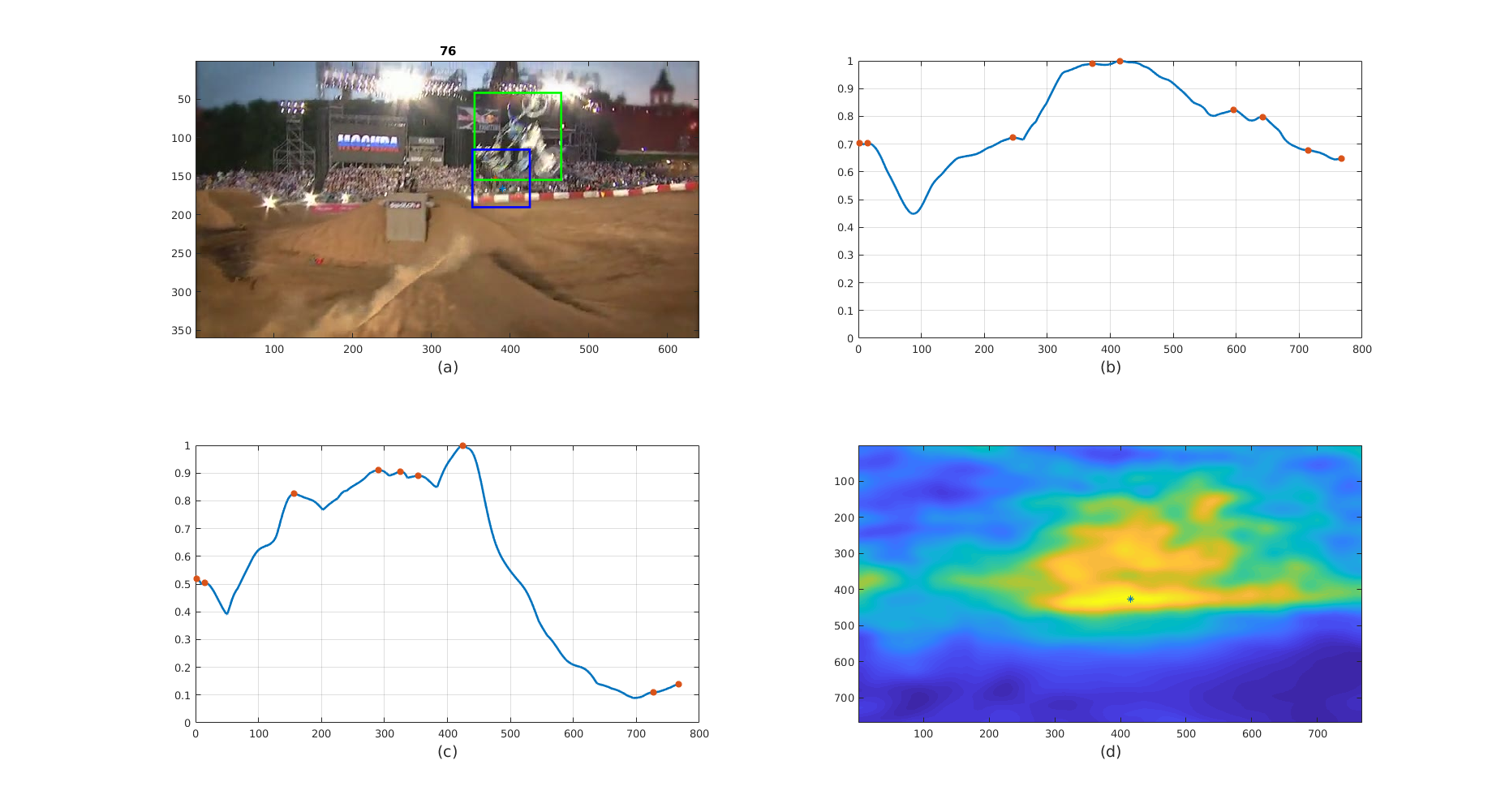}
\caption{Example of a tracking motorbike: due to the presence of multiple peaks (b-c) in the feature map (d), our HnS method rejects the peak, passes to failure mode and retrieves the motorbike in the next frame (a), instead of failing.}
\label{fig:moto}}
\end{figure}

Finally, as an explicative example, we report the very challenging case of tracking a motorbike in Figure \ref{fig:moto}. In frame 76 (Figure \ref{fig:moto}(a)) the baseline tracker is going to lose the target due to a false detection in the background. However, our HnS method finds two peaks in the feature map (Figure \ref{fig:moto}(d)) through its projection onto $yz$ and $xz$ planes (Figures \ref{fig:moto}(b) and \ref{fig:moto}(c), respectively). Therefore, HnS avoids the sampling drift by passing to failure mode, before recovering the bike position in the next frame.

\section{Conclusions and future developments}\label{sec:concl}

In this work we examined the hypothesis that in tracking-through-similarity algorithms the output heatmap that determines the object position in the next frame could be used to alert about possible sampling drift.
The experimental results confirm this hypothesis, while additional validating the use of a fast and simple classical tracker as a ``failure mode'' tracker, which would estimate the target position in the current frame allowing the tracker to recover in the next frames.
Developing a reliable recovery strategy in case of object loss is crucial for real world applications of video tracking technology where the behaviour of targets is often more complicated that what happens in benchmark videos.
Finally, the benefits from a slightly more elaborate running average method suggest that the use of a deep-learning approach (such as a RNN) has a big potential for further improvement.
This would be our main focus in the near future.


\end{document}